\documentclass{llncs}
\usepackage{graphicx}

\begin{document}

\title{Generating Synthetic Data for Text Recognition}
%
%
\author{Praveen Krishnan \and C.V. Jawahar
}
%
%
%
\institute{CVIT, IIIT Hyderabad, India\\
\email{praveen.krishnan@research.iiit.ac.in, jawahar@iiit.ac.in}}

\maketitle              

\begin{abstract}
Generating synthetic images is an art which emulates the natural process of image generation in a closest possible manner. In this work, we exploit such a framework for data generation in  handwritten domain. We render synthetic data using open source fonts and incorporate data augmentation schemes. As part of this work, we release 9M synthetic handwritten word image corpus which could be useful for training deep network architectures and advancing the performance in handwritten word spotting and recognition tasks.

\end{abstract}

\section{Introduction and Related Work}
\label{sec:intro}
Quality data~\cite{Everingham10,imagenet_cvpr09} has always played a pivotal role in the advancement of pattern recognition problems. Some of the key properties for any dataset are:- (i) a good sample distribution which can mimic the real world unseen examples, (ii) quality of annotation and (iii) scale. With the success of deep learning based methods~\cite{KrizhevskySH12,SimonyanZ14a,SzegedyLJSRAEVR15,JaderbergSVZ14}, there has been surge in newer supervised learning architectures which are ever more data hungry. These architectures have millions of parameters to learn, thereby need large amount of training data to avoid over-fitting and generalize well. In general data creation is a time consuming and expensive process which requires huge human efforts starting from the data collection phase to annotation and validation. More recently an alternative form of data generation process with minimal supervision is getting popular~\cite{JaderbergSVZ14,RozantsevLF15,ros2016synthia} which uses synthetic mechanisms to render and annotate the images in appropriate form. The simple idea of generating data synthetically allows to overcome challenges in problems where the data is inherently difficult to obtain and problems which require huge data to train such as object detection~\cite{RozantsevLF15,RematasRFT14} and semantic segmentation in 3D models~\cite{ros2016synthia}. One of the first successful use of synthetic data for document image processing is shown in~\cite{SankarJM10} which uses rendered images for the task of annotating large scale of document images. A similar scheme has been used in~\cite{Rodriguez-SerranoPLS09} for querying in handwritten collection. In~\cite{JawaharBMN09}, a scheme of synthesis using online handwritten samples using strokes is demonstrated. In this work, we address the need for large scale annotated datasets for handwritten images by generating synthetic word images with natural variations.

Recognition of handwritten (HW) data is one of oldest sought out challenges in the realms of artificial intelligence with great success in classification of handwritten characters and digits. However it is somewhat disappointing that there are no enough works in the space of offline recognition or retrieval for handwritten word images which could enable practical applications. We believe that with the availability of a large scale handwritten data which captures the inherent challenges in terms of numerous styles and distortion, one could address this issue to a large extent. 

\begin{figure}[t]
\centering
\includegraphics[height=4.5cm]{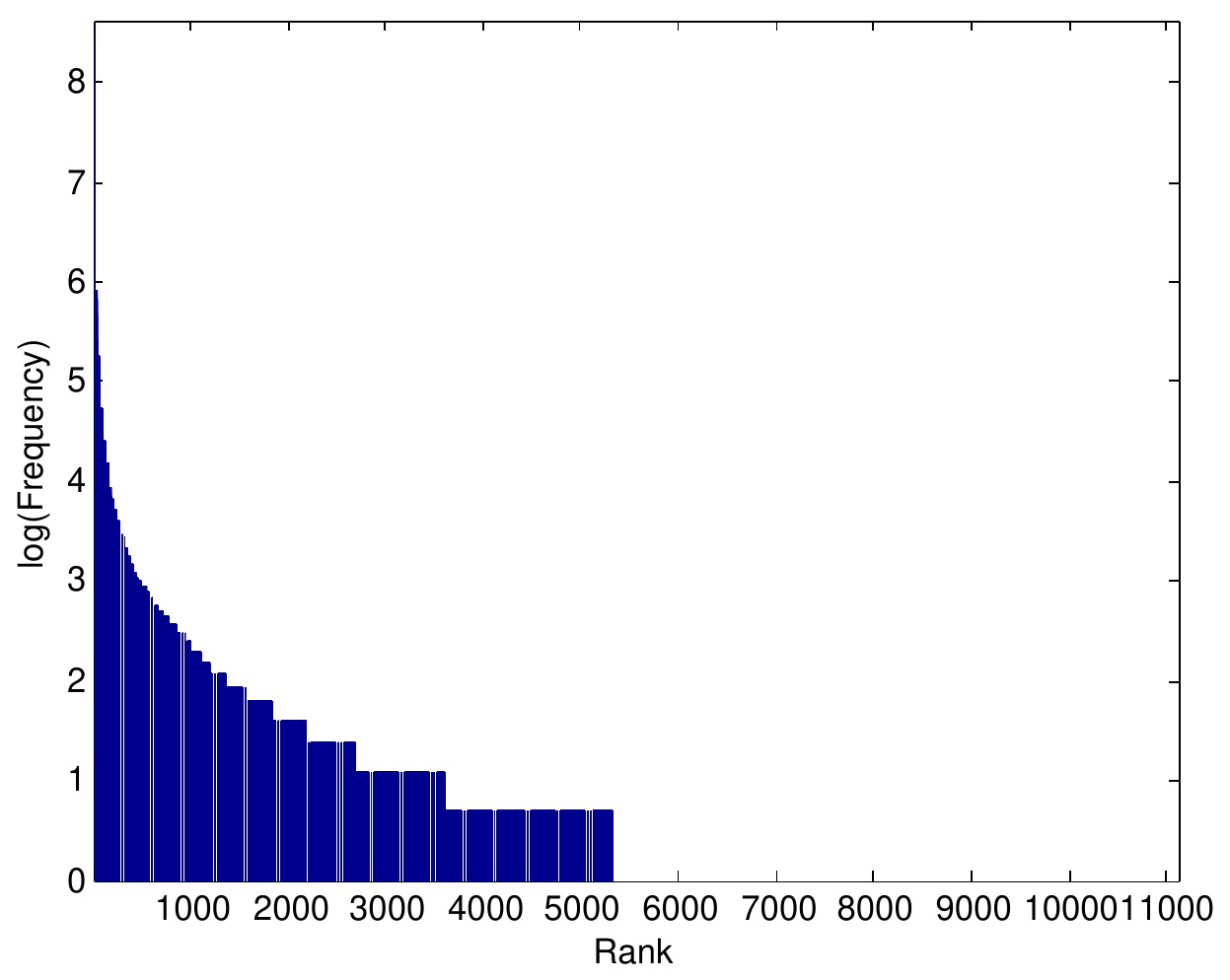}
\caption{Distribution of words in IAM dataset.}
\label{fig:iamWordsDist}
\end{figure}

Some of the popular datasets in HW domain are IAM  handwriting dataset~\cite{Marti02}, George Washington dataset~\cite{Fischer12,ManmathaHR96}, Bentham manuscripts~\cite{causer2012building}, Parzival database~\cite{Fischer12} etc. Except IAM, the remaining datasets are part of historical collection which were created by one or very few writers. In historical documents, the major challenges involves in predicting rare usages of ligatures, handling degradation of document while the style is more or less consistent. IAM is a relatively modern dataset, published in ICDAR 1999 which consists of unconstrained text written in forms by 657 writers. The vocabulary of IAM is limited by nearly 11K words whereas any normal dictionary in English language would contain more than 100K words. Fig.~\ref{fig:iamWordsDist} shows the distribution of entire words in IAM vocabulary which follows the typical Zipf law. As one can notice that out of 11K words, nearly 10.5K word classes contain fewer than 20 samples or instances. Also the majority of remaining words are stop words which are shorter in length and are less informative. The actual samples in training data is much smaller than this which limits building efficient machine learning models such as deep learning networks~\cite{KrizhevskySH12}. 

\section{Synthetic Word Image Rendering}
\label{sec:synthImage}
Creation of synthetic data for word images can approached in two ways:- (i)
rendering words using the available font
classes~\cite{JaderbergSVZ14,CamposBV09,WangWCN12} and (ii) learning latent model
parameters to separate style and content, thereby modifying the styles
aspects alone for different variation. Though the later work seems promising
with success in generating characters but synthesizing word images is still a
challenging task. In this work, we extend the former approach due to recent 
availability of handwritten fonts and manipulate the rendered images to simulate
the natural variations present in handwritten domain.

\begin{figure}[t]
\centering
\includegraphics[height=2.8cm,width=8.5cm]{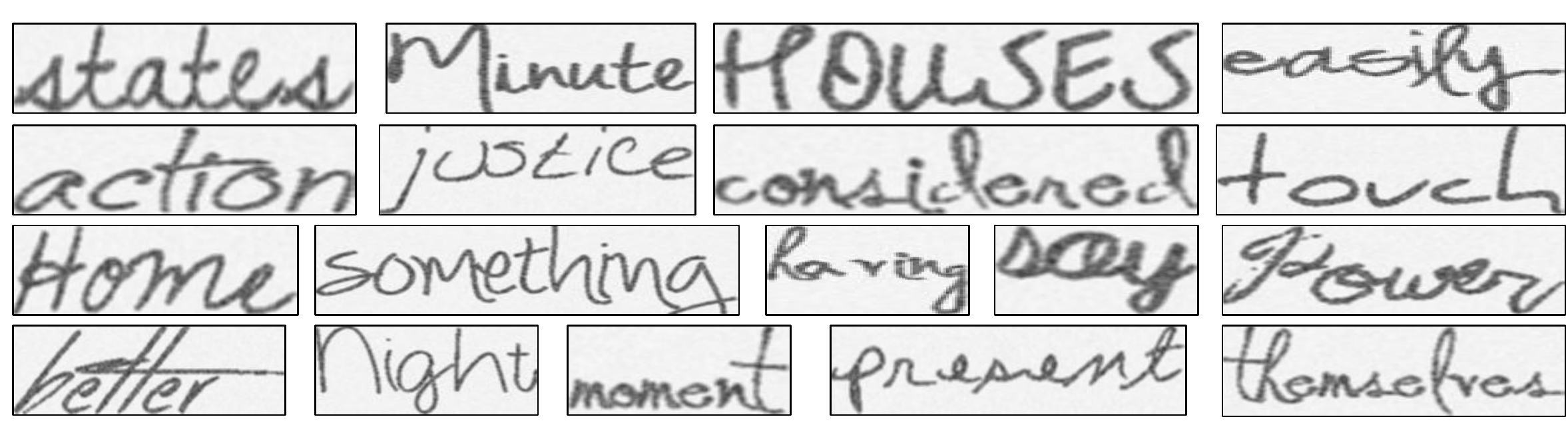}
\caption{Sample synthetic handwritten word images rendered in this work. Notice
the variability of each word image which mimics the natural writing process.}
\label{fig:varImages}
\end{figure}

\subsection{Handwritten Font Rendering}
\label{subsec:fonts}
We use publicly available handwritten fonts for our task. The vocabulary of words is chosen from a dictionary. 
For each word in vocabulary, we randomly sample a font and render\footnote{We
use ImageMagick for the rendering of word images. URL: \url{http://www.imagemagick.org/script/index.php}} its 
corresponding image. During this process, we vary the following parameters: 
(i) kerning level (inter character space), (ii) stroke width, from a defined
distribution. In order to make the pixel distribution of both foreground ($F_g$) and
background ($B_g$) pixels more natural, we sample the corresponding pixels for both
regions from a Gaussian distribution where the parameters such as mean and
standard deviation are learned from the $F_g$ and $B_g$ region of IAM dataset. Finally
a Gaussian filtering is done to smooth the rendered image.

It is well know that natural handwriting is also affected by many factors other than the inherent writing style of the writer such as (i) angle at which the writing medium is placed, (ii) speed of writing, (iii) friction between the pen and writing medium and other motor functionalities associated with hand movement. In the current work, we limit to affine transformation to capture few of such variations. We apply a random amount of rotation ($+/-5$ degrees), shear ($+/-0.5$ degrees along horizontal direction) and perform translation in terms of padding on all four sides to simulate incorrect segmentation of words. 

\subsection{IIIT-HWS dataset}
\label{subsec:hwsynth}
Inspired from~\cite{JaderbergSVZ14}, to address the lack of data in handwritten images, we release IIIT-HWS dataset comprising of nearly 9M synthetic word images rendered out of 750 publicly available handwritten fonts. We use 90K unique words as the vocabulary which is picked from a popular open source English dictionary Hunspell. For each word in the vocabulary, we randomly sample 100 fonts and render its corresponding image and follow the post processing steps as described in the previous section. Figure~\ref{fig:varImages} shows some sample rendered word images using handwritten fonts which tries to simulate actual handwritten words.

\section{Conclusion}
\label{sec:conc}
In this work, we propose a framework to render large scale synthetic data for handwritten images. We also release IIIT-HWS dataset, a 9M word image corpus for the purpose of training deep neural networks which would enable learning better models for handwritten word spotting and recognition tasks. In the current work, we have not addressed how to include the cursive property present in the word images. As a future work we plan to address this issue along with additional augmentation schemes such as elastic distortion.

\bibliographystyle{splncs}
\bibliography{ref}

\end{document}